\documentclass[a4paper,conference]{IEEEtran}

\pdfoutput=1

\usepackage[OT4,T1]{fontenc}
\usepackage[utf8]{inputenc}
\usepackage{qtree}
\usepackage{amsmath}
\usepackage{url}       
\usepackage[margin=25mm]{geometry}
\usepackage{times}

\usepackage{color}
\usepackage[dvips]{graphicx}
\usepackage{caption2}
\usepackage[outdir=./]{epstopdf}

\newsavebox{\topbox}
\newsavebox{\leftbox}
\newsavebox{\rightbox}

\newsavebox{\tmpstrikebox}
\newlength{\tmpstrikelen}

\newcommand{\strikeheight}[2]{%
\sbox{\tmpstrikebox}{#1}%
\settowidth{\tmpstrikelen}{\usebox{\tmpstrikebox}}%
\makebox[0pt][l]{\rule[#2]{\tmpstrikelen}{0.4pt}}\usebox{\tmpstrikebox}%
}
\newcommand{\strike}[1]{\strikeheight{#1}{.5ex}}
\newcommand{\ssim}{\raise.17ex\hbox{$\scriptstyle\mathtt{\sim}$}}

\IEEEoverridecommandlockouts

\title{Grammatical Case Based IS-A Relation Extraction with Boosting for Polish \thanks{The study is cofounded by the European Union from resources of the European Social Fund. Project PO KL ,,Information technologies: Research and their interdisciplinary applications'', Agreement UDA-POKL.04.01.01-00-051/10-00.}}

\author{
\IEEEauthorblockN{Paweł Łoziński, Dariusz Czerski, Mieczysław A. Kłopotek}
\IEEEauthorblockA{Institute of Computer Science\\
Polish Academy of Sciences\\
ul. Jana Kazimierza 5, 01-248 Warsaw, Poland\\
Email: \{pawel.lozinski, dariusz.czerski, mieczyslaw.klopotek\}@ipipan.waw.pl}
}

\begin{document}

\maketitle

\begin{abstract}
\boldmath
Pattern-based methods of IS-A relation extraction rely heavily on so called Hearst patterns. These are ways of
expressing instance enumerations of a class in natural language.
While these lexico-syntactic patterns prove quite useful, they may not capture all taxonomical relations expressed in
text. Therefore in this paper we describe a novel method of IS-A relation extraction from patterns, which uses
morpho-syntactical annotations along with grammatical case of noun phrases that constitute entities participating in IS-A relation.
We also describe a method for increasing the number of extracted relations that we call \emph{pseudo-subclass boosting}
which has potential application in any pattern-based relation extraction method.
Experiments were conducted on a corpus of about 0.5 billion web documents in Polish language.
\end{abstract}


%
\IEEEpeerreviewmaketitle

\section{Introduction}
\IEEEPARstart{R}{elation} extraction is a necessary step of any ontology induction or taxonomy induction task. Typically it takes as
input morpho-syntactically annotated text and produces a set of triples $(E_1, R, E_2)$, where $E_1$ and $E_2$ are
entities and $R$ is a relation in which $E_1$ and $E_2$ participate as a pair. In case of ontology induction or
information extraction in open domain (as described, e.g., in \cite{poon2010unsupervised},
\cite{Fader:2011:IRO:2145432.2145596}, \cite{Banko:2007:OIE:1625275.1625705}, \cite{Etzioni:2011:OIE:2283396.2283398})
no restrictions are imposed on $R$.
There are many types of relations that can be extracted this way, such as \emph{quality}, \emph{part} or \emph{behavior}
\cite{Barbu20153501}. In case of taxonomy induction the main interest is in the IS-A (hyponym-hypernym) relation.
Approaches to IS-A extraction described in literature rely on evidence from pattern extraction and statistical information (cf. \cite{probase}, \cite{fountain2012taxonomy},
\cite{cimiano2005learning}). Pattern-based methods rely heavily on so called Hearst patterns, first described in
\cite{Hearst:1992:AAH:992133.992154}. These are ways of expressing instance enumerations of a class in natural language.
Typical forms are ,,c \emph{such as} i1, i2 or i3'' or ,,c, \emph{for example} i1, i2 or i3''. Terms extracted with such
patterns may serve as input for elaborate taxonomy and ontology construction methods as, e.g., \cite{Kozareva2014}.
 While these lexico-syntactic patterns prove quite useful, they may not capture all taxonomical relations expressed in
text. Therefore in  
this paper we
describe a novel method of IS-A relation extraction from patterns, which uses morpho-syntactical annotations along with
grammatical case of noun phrases that constitute entities participating in IS-A relation.
The method is unsupervised, as it is based on hand-crafted patterns, dictionary filtering and manually adjusted
support level. Precision of this method, understood as the ratio of correct extracted IS-A relations to all extracted
relations is estimated using manual scoring of about 110 relations randomly selected from the method's output.  
Based on an internet corpus of documents, the method produces a big number of IS-A relations. Most of them (roughly
90\%) occur only once in the corpus introducing a high level of noise.
We
show in conducted experiments that even for a slight increase  of support (given as a number of occurrences), the
estimated precision of this method increases strongly.
We also describe a new method for increasing the number of extracted relations for any support level bigger than 1. 
The method is based on very simple heuristic for detection of hyponymy between class part of extracted relations, thus
we call it \emph{pseudo-subclass boosting} (\emph{psc} in short).
It is worth mentioning that this boosting approach can be applied in any pattern-based relation extraction method. 
Experiments were conducted on a
corpus of about 0.5 billion web documents in Polish language crawled in NEKST project (\url{http://www.nekst.pl}) and maintained up to
date. These include primarily HTML documents, but also other formats found on websites like PDFs and DOCs. In order to
process such high volume of data it was implemented using MapReduce framework \cite{Dean:2008:MSD:1327452.1327492} implemented in Apache Hadoop project (\url{http://hadoop.apache.org})
and Hive (\url{http://hive.apache.org}).
All examples mentioned in the article are real data, taken from working instance of NEKST system.

\section{Our approach}\label{s:our_approach}
It is known that languages that have inflection and free word order are much harder for automatic 
analysis\footnote{See e.g. \cite{xu_et_al_2002b} for problems with relation mining in German, in which the word order
is much less free than in Polish; note that they use an initial lexicon while we do start from scratch when extracting
relations.} than, e.g., English.
As pointed out in \cite[pp. 100]{nivre2007maltparser}, free word order implies non-projective grammar. It is shown in
\cite{Mcdonald06onlinelearning} and \cite{mcDonald2007} that dependency parsing for non-projective grammars is
NP-hard, apart from a very narrow subclass called edge-factored grammars. This challenge is addressed, among others, by
transition-based dependency parsing \cite{KuhlmannTTF10} used in the preprocessing step for the algorithm described in
this paper. 
We argue that inflection in a language is not only a drawback but can also be a great advantage.
Typical constructs that express the hypernymy relation explicitly in Polish language are:
\\
\begin{equation}\label{c:nominative}
NP_1^{Nom} \text{~to~} NP_2^{Nom},
\end{equation}
\begin{equation}\label{c:instrumental_1}
NP_1^{Nom} \text{~jest~} NP_2^{Abl}.
\end{equation}
\\
Both of them are a way of saying $NP_1$ \emph{is} $NP_2$ and in both cases noun phrase $NP_1$ is expressed in
nominative. They differ in grammatical case of $NP_2$, where in the first construct we have nominative and in the
second: instrumental. The second pattern has its equivalent for past tense:
\\
\begin{equation}\label{c:instrumental_2}
NP_1^{Nom} \text{~był/była/było~} NP_2^{Abl}.
\end{equation}
\\
Obviously in case of past tense construction it is possible that IS-A relation no longer holds\footnote{The relation was
valid in the past only}.
The problem exists to a lesser extent also in present tense, which for example can be a consequence of outdated web documents.
Assessment of correctness with respect to a given point in time is, in our opinion, a research direction of its own, thus it is out of scope of this paper.

As will be shown later, combination of word and grammatical case pattern allows for relation
extraction with quite high precision. It is possible partially thanks to the fact that instrumental case in Polish language
is \emph{regular} for nouns and has unique suffixes shown is Table \ref{t:instrumental} (after \cite[pp. 145,
148]{nagorko2007}). This makes automatic analysis of sentence tokens easy for this case.

\begin{table}
\begin{center}
\begin{tabular}{|l|c|c|c|}
\hline
                   & \textbf{masculine} & \textbf{neuter} & \textbf{feminine} \\
\hline
\textbf{singular} & \multicolumn{2}{|c|}{-em} & -ą \\ 
\hline
\textbf{plural}  & \multicolumn{3}{|c|}{-ami (-mi)} \\
\hline
\end{tabular}
\end{center}
\caption{Suffixes in instrumental case for Polish}
\label{t:instrumental}
\end{table}

We propose a rule-based approach for IS-A relation extraction with the following procedure:
\begin{itemize}
  \item run each sentence in corpus through POS-tagger and dependency parser,
  \item select dependency trees with promising structure,
  \item apply dictionary filtering for the head of $NP_2$,
  \item apply a set of construction rules to dependency tree in order to build instance name out of $NP_1$ and
  class name out of $NP_2$,
  \item apply a set of filtering rules.
\end{itemize}
This method is additionally extended with a technique that we call \emph{pseudo-subclass boosting} which increases the
number of extracted relations.

It is worth noting that \emph{automatic} detection of IS-A patterns is possible. Experiments described in
\cite{ilprints665} show that hand-crafted ontologies like WordNet can be used successfully as a training set for such
pattern discovery task. However, our problem setting differs from that research significantly. Apart from the already
mentioned inflection challenge and free word order language, our corpus consists of about 11 billion sentences, which is
four orders of magnitude more than the Reuters corpus used in \cite{ilprints665} and imposes efficiency limitations.
On the other hand, the gain in size comes at the price of quality -- Internet documents tend to have much more noisy
content than printed journal articles. We have no knowledge of any research on IS-A patterns detection in similar
setting (that is web-scale), which leads us to first tackle a more realistic problem of extracting IS-A relations with
\emph{known} patterns.
Nevertheless, this is a task worth trying given experience gained from research reported here.

\subsection{POS tagging and dependency parsing}

For part-of-speech tagging we use the Apache OpenNLP (\url{http://opennlp.apache.org}) tagger trained with Maximum
Entropy classifier on NKJP \cite{nkjp} corpus. Additionally, for known words, we optimized the tag disambiguation process by narrowing tags that can be chosen by information taken from the
PoliMorf dictionary \cite{WOLISKI12.263}. For Polish language, whose tagset contains around 1000 tags \cite{prze:04ce},
this simple optimization gives an improvement of tagging in terms of accuracy and processing speed at the same time. To
give an example, the word \emph{artykułów} (inflected form of the word \emph{article}) has only two possible tags
\texttt{subst:pl:gen:m3} and \texttt{subst:pl:gen:p3}. Using this knowledge in OpenNLP tagger reduces search space for
this word 500 times. 
Dependency parsing is based on MaltParser framework
\cite{NLE:1012768} trained on Polish Dependency Bank that consists of 8030 sentences \cite{wroblewska2012PDB}. To obtain
high processing speed (essential for such large volume of text data) the liblinear classification model has been used.

\subsection{Promising dependency tree structure selection}
\begin{figure}[htp]
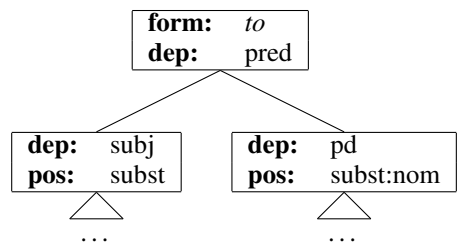

  \Tree [.\usebox{\topbox} \qroof{\ldots}.\usebox{\leftbox} \qroof{\ldots}.\usebox{\rightbox} ] \hskip 0.3in%
  \caption{Dependency tree structure for construct (\ref{c:nominative})}
  \label{struct_1}
\end{figure}

\begin{figure}[htp]
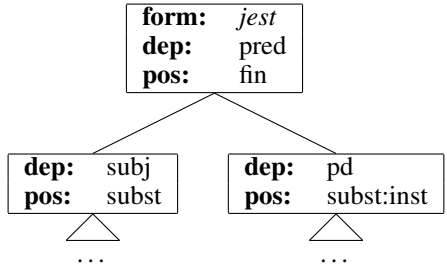

  \Tree [.\usebox{\topbox} \qroof{\ldots}.\usebox{\leftbox} \qroof{\ldots}.\usebox{\rightbox} ]%
  \caption{Dependency tree structure for construct (\ref{c:instrumental_1})}
  \label{struct_2}
\end{figure}

\begin{figure}[htp]
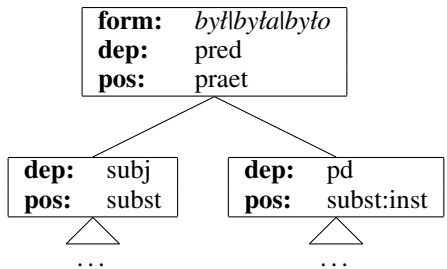

  \Tree [.\usebox{\topbox} \qroof{\ldots}.\usebox{\leftbox} \qroof{\ldots}.\usebox{\rightbox} ]
  \caption{Dependency tree structure for construct (\ref{c:instrumental_2})}
  \label{struct_3}
\end{figure}

By \emph{promising} structure of a dependency tree we mean one that matches any of the patterns depicted in Figures
\ref{struct_1}, \ref{struct_2} and \ref{struct_3}, where \textbf{form}, \textbf{dep} and \textbf{pos} mean: token form, dependency relation type (as described in
\cite{wroblewska2012PDB}) and part-of-speech tag (as described in \cite{nkjp}) respectively. 

\qroofy=2
\qroofx=2
\setbox\topbox=\hbox{
	\begin{tabular}{|ll|}
		\hline
		\textbf{form:} & \emph{to} \\
		\textbf{dep:}  & pred \\
		\hline
	\end{tabular} 
}
\setbox\leftbox=\hbox{
	\begin{tabular}{|ll|}
		\hline
		\textbf{dep:}  & subj \\
		\textbf{pos:}  & subst \\
		\hline
	\end{tabular} 
}
\setbox\rightbox=\hbox{
	\begin{tabular}{|ll|}
		\hline
		\textbf{dep:}  & pd \\
		\textbf{pos:}  & subst:nom \\
		\hline
	\end{tabular} 
}

\setbox\topbox=\hbox{
	\begin{tabular}{|ll|}
		\hline
		\textbf{form:} & \emph{jest} \\
		\textbf{dep:}  & pred \\
		\textbf{pos:}  & fin \\
		\hline
	\end{tabular} 
}
\setbox\leftbox=\hbox{
	\begin{tabular}{|ll|}
		\hline
		\textbf{dep:}  & subj \\
		\textbf{pos:}  & subst \\
		\hline
	\end{tabular} 
}
\setbox\rightbox=\hbox{
	\begin{tabular}{|ll|}
		\hline
		\textbf{dep:}  & pd \\
		\textbf{pos:}  & subst:inst \\
		\hline
	\end{tabular} 
}


In both nominative and instrumental case, the base structure has a predicate word with outgoing dependency arcs to two 
other words with subjective and predicative complement relation type. 
The difference between structure \ref{struct_1}, \ref{struct_2} and \ref{struct_3} is in the 
grammatical case of the predicative complement and part of speech of the predicate.
Our intuition is that selected structures are natural sources of IS-A relation. This claim is supported by the
estimated precision obtained in conducted experiments.

\qtreecentertrue
\setbox\topbox=\hbox{
    \begin{tabular}{|ll|}
        \hline
        \textbf{form:} & \emph{był}|\emph{była}|\emph{było} \\
        \textbf{dep:}  & pred \\
        \textbf{pos:}  & praet \\
        \hline
    \end{tabular} 
}
\setbox\leftbox=\hbox{
    \begin{tabular}{|ll|}
        \hline
        \textbf{dep:}  & subj \\
        \textbf{pos:}  & subst \\
        \hline
    \end{tabular} 
}
\setbox\rightbox=\hbox{
    \begin{tabular}{|ll|}
        \hline
        \textbf{dep:}  & pd \\
        \textbf{pos:}  & subst:inst \\
        \hline
    \end{tabular} 
}

Figure
\ref{f:treePaternMatchExample} illustrates an example of sentence that matches pattern \ref{struct_2}, parsed with our dependency
parser and printed in CoNLL \cite{Buchholz:2006:CST:1596276.1596305} format.
\begin{figure*}[htp]
\begin{center}
	\begin{tabular}{llllllll}
		1 & Golden & golden & subst & subst & sg:nom:m3 & 3 & subj \\	
		2 & retriever & retriever & subst & subst & sg:nom:m2 & 1 & app \\	
		3 & jest & być & fin & fin & sg:ter:imperf & 0 & pred \\	
		4 & psem & pies & subst & subst & sg:inst:m2 & 3 & pd \\	
		5 & myśliwskim & myśliwski & adj & adj & sg:loc:m3:pos & 4 & adjunct \\	
		6 & . & . & interp & interp & \_ & 3 & punct
	\end{tabular}
	\caption{Tree pattern match example in CoNLL format for the Polish sentence "Golden retriever jest psem myśliwskim" (\emph{Golden retriever is a hunting dog})}
	\label{f:treePaternMatchExample}
\end{center}
\end{figure*}
It is worth noting that in this case the part-of-speech tagger made an error in assigning a case to the adjective \emph{myśliwski}
(hunting), where instrumental instead of locative should appear. This may happen because singular masculine adjective suffixes for instrumental (as
noted in \cite[p. 160]{nagorko2007}) are not unique as with nouns. That's why in our analysis we focus only on the grammatical
case of the head of noun phrase and assume the same case for its dependent adjective tokens. This assumption is
justified by the fact, that for Polish language agreement exists between noun and adjective in a noun phrase
\cite[p. 174]{saloni2011}.
POS tag in the example is repeated twice because CoNLL format specifies CPOSTAG and POSTAG allowing for coarse-grained and fine-grained part-of-speech tagsets which are the same for Polish language. The following steps illustrate how pattern \ref{struct_2} applies to the example sentence from figure \ref{f:treePaternMatchExample}:
\begin{itemize}
  \item find a root word of the sentence (\emph{jest} in our case), and check its dependency relation (must be \emph{pred}) and a POS tag (must be \emph{fin}),  
  \item if the root word has two descendants, then test if:
  \begin{itemize}
  \item its left descendant (\emph{golden}) has correct dependency relation (must be \emph{subj}) and a POS tag (must be \emph{subst}),
  \item its right descendant (\emph{pies}) has correct dependency relation (must be \emph{pd}) and a POS tag (must be \emph{subst:inst}),
\end{itemize}
\item if all requirements are fulfilled, the sentence is moved to the phase of dictionary filtering
(section \ref{ss:dict-filter}) and instance and class name construction (section \ref{ss:construction_rules}).
\end{itemize}

Given a sentence whose dependency tree matches one of above-mentioned patterns, we construct $NP_1$ from its left sub-tree 
and $NP_2$ from its right sub-tree. Head (or root) of left and right sub-tree will be denoted $N_1^H$ and $N_2^H$
respectively.

\subsection{Dictionary filtering for the head of $NP_2$}\label{ss:dict-filter}
Preliminary experiments showed that many of sentences matching constructs (\ref{c:nominative}) and (\ref{c:instrumental_1})
contain very general, ambiguous nouns in $NP_2$ like \emph{problem}, \emph{aspect}, \emph{element} or \emph{outcome}.
Those nouns cannot be considered proper classes in the sense of IS-A relation, rather they are catch-all phrases used to
express various thoughts about what is contained in $NP_1$.

We eliminated those nouns by manually evaluating a random sample of about 1000 experiment results and creating a
dictionary of such meaningless ,,classes''. In this step of our extraction procedure we filter extractions with this
dictionary. This process was repeated in three iterations. Size of the dictionary started with 95 catch-all phrases
increased by 50, and 20 reaching the level of about 170.

\subsection{Construction rules for $NP_1$ and $NP_2$}\label{ss:construction_rules}
We construct both instance name (from $NP_1$) and class name (from $NP_2$) out of lemmatized tokens. The first step is to
serialize tokens present in both dependency sub-trees with operators \emph{leftOffspring} and \emph{rightOffspring}, which operate as follows:
\begin{enumerate}
  \item put all nodes of dependency sub-tree in a list $L$,
  \item sort $L$ by CoNLL token id descending (for \emph{leftOffspring} operator)/ascending (for
  \emph{rightOffspring} operator),
  \item find index $i_H$ of sub-tree head in $L$,
  \item create sub-list $L'$ from $i_H$ to the first occurrence of interpunction or end of $L$,
  \item in case of \emph{leftOffspring}: sort $L'$ by CoNLL token id ascending,
  \item concatenate lemmas of tokens in $L'$ and return. 
\end{enumerate}
Computational complexity of this algorithm is $O(n)$, where $n$ is the sentence length. Actual sorting of tokens in case
of steps 2. and 5. is not necessary and was introduced to simplify the description\footnote{It unifies the procedure for left and right part of the sentence.}.
 Boundaries detection of instance name is quite simple because it is typically directly defined by
left sub-tree of all considered dependency structures (fig. \ref{struct_1}, \ref{struct_2} and \ref{struct_3}). Therefore it is constructed as
concatenation:
\\
\begin{equation*}
\mathit{leftOffspring}(N_1^H) + N_1^H + \mathit{rightOffspring}(N_1^H)
\end{equation*}
\\
Creation of class name is more complicated as it is often preceded by degrees of comparison and followed
by the rest of the sentence which may be loosely coupled with the class itself. Consider the following sentences:
\begin{quote}
Trójmorski Wierch jest jedyną polską górą, z której spływają wody aż do trzech mórz.
\end{quote}
(\emph{Trójmorski Wierch is the only Polish mountain, from which waters flow to as many as three seas.})
\begin{quote}
Korona norweska to waluta oznaczana międzynarodowym kodem – NOK.
\end{quote}
(\emph{Norwegian krone is a currency marked with the international code -- NOK.})

In the first example, the word \emph{jedyną} (the only) cannot be considered as part of class name. Likewise, anything
that comes after word \emph{waluta} (currency) in the second example is merely a description of Norwegian krone, not
part of a class name. To address such issues construction rules for class name simply omit the output of
\emph{leftOffspring} operator and truncate \emph{rightOffspring} output: it is iterated from left to
right only as long as the tokens have POS tag from set \{adj, subst, ger\} \emph{and} dependency type from set
\{adjunct, app, conjunct, obj\}. So the class name results from concatenating:
\\
\begin{equation*}
N_2^H + \mathit{truncate(rightOffspring}(N_2^H))
\end{equation*}

This forces extraction of shorter phrases, which increases the probability of observing a given instance-class pair
more than once. As we show in section \ref{s:ex}, this highly influences the precision of the method.
Extraction results for above examples are:
\emph{Trójmorski Wierch} \texttt{IS-A} \emph{góra} \lbrack\emph{Trójmorski Wierch IS-A mountain}\rbrack\ and \emph{Korona norweska} \texttt{IS-A} \emph{waluta}  \lbrack\emph{Norvegian crown IS-A
currency}\rbrack, while from such sentence:
\begin{quote}
Narodowy Bank Belgijski jest bankiem centralnym od 1850 roku.\footnote{Belgian National Bank is the central bank
since 1850.}
\end{quote}
we acquire \emph{Narodowy Bank Belgijski} \texttt{IS-A} \emph{bank centralny} \lbrack\emph{Belgian National Bank IS-A central bank}\rbrack.   

\subsection{Final filtering rules}\label{final_filtering}

It is common that $NP_1$ contains reference to earlier parts of text. Two types of such reference can be
distinguished:
\begin{enumerate}
  \item explicit:
  \begin{quote}
      Ten wikipedysta jest numizmatykiem.\footnote{This wikipedian is a numismatist.}
  \end{quote}
  \item implicit:
  \begin{quote}
      Pisarka jest członkiem Związku Pisarzy Białorusi.\footnote{The writer is a member of Union of Belarus Writers.}
  \end{quote}
\end{enumerate}
In both cases $NP_1$ typically contains a class of referenced entity, not the entity itself
which leads to erroneous extractions. As long as this reference is explicit, we filter such cases with a dictionary of
referencing words (pronouns and textual references like \emph{above-mentioned}). The case where reference is
implicit is much harder, and at this point left for further research, as described later in section \ref{s:future}.

\subsection{Pseudo-subclass (psc) boosting}
Our experiments showed that the number of extracted relations drops significantly with increase of support level $t$. To compensate
this loss we designed a boosting method that is based on the following intuition: if \emph{I} \texttt{IS-A} \emph{C} and
\emph{I} \texttt{IS-A} \emph{C'} are extracted relations and \emph{C} is a substring of \emph{C'}, then there is high
chance that \emph{C'} is a way of describing \emph{I} more precisely than \emph{C}, i.e., \emph{C'} is a pseudo-subclass
of \emph{C}. If so, we can boost our confidence in the fact that \emph{I} \texttt{IS-A} \emph{C} is properly extracted.
To give an example:
\begin{quote}
Kraków to najchętniej odwiedzane miasto przez turystów w Polsce.
Kraków -- dawna stolica Polaków jest miastem magicznym.\footnote{Cracow is the most visited city by tourists in Poland.
Cracow -- the former capital of the Poles is a magical city.}
\end{quote}

Above two sentences allow for boosting confidence in extraction \emph{Kraków} IS-A \emph{miasto} (\emph{Cracow} IS-A
\emph{city}). 
From the first sentence we get the relation \emph{Kraków} IS-A \emph{miasto} and from the second \emph{Kraków} IS-A \emph{miasto magiczne} (\emph{Cracow}
IS-A \emph{magic city}). As "miasto magiczne" is a superstring of "miasto", the second sentence supports the first extracted relation.
In general, to detect class/pseudo-subclass matches for each extraction \emph{R = I} \texttt{IS-A} \emph{C} we
generate a list \emph{L} of
\begin{itemize}
  \item prefix lists of tokens from \emph{C},
  \item suffix lists of tokens from \emph{C} that don't include leading adjectives. 
\end{itemize}
In Map phase of MapReduce job, we emit the pair \emph{(I, C)} with \emph{R}'s occurrence count and pairs \emph{(I, c)} (with the same
count) for each $c \in L$. Reduce phase aggregates our data by matched pairs and here we acquire knowledge about
pseudo-subclasses' occurrence count and type of constructs they were discovered in. Figure \ref{psc_example} illustrates
a more elaborate case of pseudo-subclass boosting. Each numbered row represents a relation \emph{mukowiscydoza}
IS-A \emph{\ldots} extracted from text. Row 13 is an example of suffix list boosting with \emph{wieloukładowa} being an
adjective removed at the stage of creating list \emph{L}. Rows 2-12 boost relation \emph{mukowiscydoza} IS-A
\emph{choroba}, additionally rows 4-7 boost \emph{mukowiscydoza} IS-A \emph{choroba genetyczna}, etc.
\begin{figure*}
\begin{verbatim}
mukowiscydoza (cystic fibrosis) IS-A
 1. choroba (disease)
 2.   choroba dziedziczna (hereditary disease)
 3.   choroba genetyczna (genetic disease)
 4.     choroba genetyczna ludzi rasy białej 
        (genetic disease of white race people)
 5.     choroba genetyczna ogólnoustrojowa (systemic genetic disease)
 6.     choroba genetyczna rasy białej (genetic disease of white race)
 7.     choroba genetyczna układu pokarmowego 
        (genetic disease of the digestive system)
 8.   choroba monogenowa (monogenic disease)
 9.   choroba nieuleczalna (incurable disease)
10.   choroba przewlekła (chronic disease)
11.   choroba wielonarządowa (multiorgan disease)
12.   choroba wieloukładowa (multisystem disease)
13.   wieloukładowa choroba (multisystem disease)
14.     wieloukładowa choroba monogenowa 
        (multisystem monogenic disease)
15. przyczyna wykonywania (cause of performing)
16.   przyczyna wykonywania przeszczepu płuca 
      (cause of performing lung transplant)
17. schorzenie (disease - synonym)
18.   schorzenie genetyczne (genetic disease - synonym)
\end{verbatim}
\caption{Tree representation of pseudo-subclass boosting.}
\label{psc_example}
\end{figure*}


\section{Experiments}\label{s:ex}
Experiments were conducted on a corpus of about 0.5 billion web documents in Polish language with roughly 11 billion
sentences. Tables \ref{t:results},  \ref{t:results_psc} and  \ref{t:results_hearst} present the results of passing the
entire collection through the algorithm described in section \ref{s:our_approach}.

Method evaluation was conducted for four levels of the value of $t$, which, as earlier described, is the minimal IS-A
relation occurrence count acceptance threshold.
Precision evaluation was based on manual scoring of about 110 randomly selected relations from given experiment's
results. Estimated precision was calculated by the formula \ref{e:prec}.
\begin{equation}\label{e:prec}
\hat{Pr} = \frac{TP}{TP+FP}
\end{equation} 
where $TP$ is the number of relations scored as correct and $FP$ is the number of relations scored as erroneous.

Tables \ref{t:results}, \ref{t:results_psc} and \ref{t:results2} show results of these experiments. Column \emph{nom} contains number of unique IS-A
relations extracted only from nominative construct, \emph{inst} is the number of unique relations only from instrumental
constructs, \emph{nom$\cap$inst} refers to count of relations extracted from nominatives and instrumentals. 
Table \ref{t:results_psc} refers to the number of relations that were
additionally accepted only thanks to pseudo-subclass boosting which helped to observe a given relation more than $t$
times or with both grammar cases.

Total number of extracted IS-A relations, for either nominative or instrumental construction, is slightly above 4 milion
(table \ref{t:results}).
Increase of support level results in drop of accepted relations (up to 1 order of magnitude between consecutive levels).
Final count of relations (for $t=4$) does not exceed 90000, which is almost 2 orders of magnitude lower than the total.            

Pseudo-subclass boosting method allows to extract around 86000 more relations at support level $2$.
Nominal number of additional relations decreases for higher support levels, but increases in terms of relative gain (as shown in the last column of table \ref{t:results_psc}).   

Estimated precision of our method is 61\% at the lowest support level, and achieves 87\% for level $4$ (table \ref{t:results2}).
Increasing the number of accepted relations with pseudo-subclass boosting comes at the cost of lower estimated precision.
At support level $2$ this loss is 1\%, but for $3$ and $4$ jumps to several percent.     
Estimated precision of our method, equipped with pseudo-subclass boosting, increases with the increase of $t$, saturating at the level of about 80\%.

\begin{table*}
\centering
\begin{tabular}{l|l|l|l|l}
          & \textbf{nom} 
          & \textbf{inst} 
          & \textbf{nom$\cap$inst} 
          & \textbf{total}\\
  \hline                                                                               
  $t = 1$ & 1647500 & 2380021 & 39865 & 4027521 \\
  $t = 2$ & 138877  & 264764  & 9895  & 403641 \\
  $t = 3$ & 52430   & 100320  & 4938  & 152750 \\
  $t = 4$ & 29210   & 55232   & 3154  & 84442 \\
\end{tabular}
\caption{Number of extracted relations for different values of manually adjusted acceptance support levels $t$.}
\label{t:results}
\end{table*}
\begin{table*}
\centering
\begin{tabular}{l|l|l|l|l|l}
          & \textbf{nom} 
          & \textbf{inst} 
          & \textbf{nom$\cap$inst} 
          & \textbf{total}
          & \textbf{psc gain}\\
  \hline                                                                               
  $t = 1$ & 0     & 0     & 0    & 0      & 0\% \\
  $t = 2$ & 24335 & 61244 & 2931 & 85579 & 21.20\% \\
  $t = 3$ & 13122 & 38004 & 2116 & 51126 & 33.47\% \\
  $t = 4$ & 8726  & 26702 & 1521 & 35428  & 41.95\% \\
\end{tabular}
\caption{Number of additional relations extracted thanks to pseudo-subclass boosting (for different values of
support level $t$).}
\label{t:results_psc}
\end{table*}

\begin{table*}
\centering
\begin{tabular}{l|l|l|l|l}
  $t$       & 1    & 2    & 3    & 4 \\
  \hline                                                                               
  precision without psc & 0.61 & 0.71 & 0.87 & 0.87 \\
  precision with psc & 0.61 & 0.72 & 0.79 & 0.81 \\  
\end{tabular}
\caption{Estimated precision ($\hat{Pr}$ -- see equation \ref{e:prec}) of extraction for different acceptance support levels.}
\label{t:results2}
\end{table*}
Experiments were performed on a cluster of 70 machines with total of 980 CPU cores and 4.375TB of RAM. Total processing
time of raw web documents: lemmatization, POS tagging, dependency parsing and IS-A relation extraction was under 24
hours.

\section{Relation to Hearst patterns}
In order to compare our method with the most popular approach, we implemented Hearst patterns extraction algorithm as
follows:
\begin{itemize}
  \item Detect enumeration phrase $R$ (one of ,,\emph{taki jak}'', ,,\emph{taki jak na
  przykład}'', ,,\emph{taki jak np.}'' which are special cases of phrase ``\emph{such as}'' in English) in a sentence,
  based on lexical constructions proposed in \cite{Hearst:1992:AAH:992133.992154}.
  \item Check if words from $R$ to the end of the sentence form a comma separated list of phrases (with the last element
  optionally separated by conjunction: ,,\emph{i}'' or ,,\emph{oraz}''). The list is assumed to represent instances of a class.
  \item Detect the class name in words left to $R$ with a Conditional Random Field model \cite{Lafferty:2001:CRF:645530.655813}. Words in this part of
  sentence are labeled with either ,,1'' or ,,0''. The sequence of ,,1'' nearest to $R$ is assumed to represent the class. 
  The model was trained on manually annotated set of around 600 sentences. Its precision calculated on 10-fold cross validation is 93.89\%. 
\end{itemize}

Table \ref{t:results_hearst} shows the number of extracted Hearst patterns, their estimated precision and overlap between
this method and our approach (percentage values in brackets are calculated relative to the number of Hearst patterns-based extractions). 
Estimated precision is substantially lower (from 14\% to 29\%). The overlap varies from 0.57\% to 1.02\% for nominative
scheme and from 1.19\% to 2.65\% for instrumental. Relations detected in all three methods constitute from 0.25\% to 0.58\% of 
relations extracted with the basic method.  
This suggests that our method allows for extraction of new relations, not expressed in language constructs described by
Hearst, with even higher precision.

\begin{table*}
\centering
\begin{tabular}{l|l|l|l|l|l}
          & \textbf{hrst}
          & $\hat{Pr}$
          & \textbf{nom$\cap$hrst} 
          & \textbf{inst$\cap$hrst} 
          & \textbf{nom$\cap$inst$\cap$hrst} \\
  \hline                                                                               
  $t = 1$ & 4007927 & 0.47 & 23044 (0.57\%) & 47953 (1.19\%) & 10222 (0.25\%) \\
  $t = 2$ & 781419  & 0.56 & 6492 (0.83\%)  & 15567 (1.99\%) & 3434 (0.44\%) \\
  $t = 3$ & 356873  & 0.58 & 3488 (0.98\%)  & 8728  (2.45\%) & 1899 (0.53\%) \\
  $t = 4$ & 224200  & 0.62 & 2295 (1.02\%)  & 5939  (2.65\%) & 1298 (0.58\%) \\
\end{tabular}
\caption{Number and estimated precision ($\hat{Pr}$ -- see equation \ref{e:prec}) of relations extracted with Hearst patterns for different values of manually
adjusted acceptance support levels $t$.}
\label{t:results_hearst}
\end{table*}

\section{Discussion}

Experiments lead to interesting conclusions. Firstly, there is little intersection between IS-A relations extracted by
the three methods: Hearst traditional method and our methods, one based on nominative, the other based on instrumental case.
The IS-A relation space seems too sparse for such methods to produce overlapping results. Nominative construction produces less
relations than instrumental, which presumably is a consequence of the fact that this construct is only applicable for present tense. 
Decrease in total
extractions count is much bigger going from support level 1 to 2 (9.98 times) than when in other cases ($2 \rightarrow
3$:
\ssim2.64 times, $3 \rightarrow 4$: \ssim1.81 times). 
It can be connected to the natural model of language, where distribution of word frequencies has power law probability distribution \cite{Manning:1999:FSN:311445}.  
There is a lot of particular, domain specific taxonomical information that is infrequent in textual resources accessible on the Internet.
On the other hand more common knowledge that can be found multiple times in text is substantially less frequent.  

Of course pseudo-subclasses don't give any boost when $t = 1$ and do not affect precision, because we simply accept
everything that passes the final filtering rules.
In other cases psc increases the number of extractions significantly (the higher $t$ the better), although not as much
as to eliminate the effect of increased $t$.
This boosting method is very beneficial for support level $2$ as it increases extractions count by 23\% with no observable
loss in precision (see Table \ref{t:results2}). For $t = 3$ and $t = 4$ the gain in extractions count comes at the price
of significantly lower precision.

Analysis of false-positive extractions reveal several types of errors made by this method:
\begin{enumerate}
  \item Implicit reference -- which leads to errors like 
  \begin{itemize}
      \item \emph{autor} \texttt{IS-A} \emph{dyrektor jednostki} (\emph{author} \texttt{IS-A} \emph{director of the unit}),
      \item \emph{sobota} \texttt{IS-A} \emph{dzień koncertu głównego} (\emph{Saturday} \texttt{IS-A} \emph{main concert day}).
  \end{itemize}
  \item Wrong decision about phrase begin/ending point%
  		\footnote{Missing parts are added in brackets, unwanted parts are striked out.}:
  		\begin{itemize}
		  \item \emph{trening funkcjonalny} \texttt{IS-A} \emph{rodzaj (\ldots{}czego?)} 
  				(\emph{functional training} \texttt{IS-A} \emph{kind (\ldots{}of what?)}),
		  \item \emph{\strike{zdecydowana większość} kandydatów do Parlamentu} \texttt{IS-A} \emph{członek określonej partii politycznej} 
			    (\strike{vast majority of} \emph{candidates to Parliament} \texttt{IS-A} \emph{member of a particular political party}).
		\end{itemize} 
  \item Ever growing dictionary mentioned in section \ref{ss:dict-filter}. After each iteration of catch-all phrases
  eliminations new such phrases emerge in result samples. Above-mentioned experiments revealed such false-positive
  classes as:
  \emph{result}, \emph{an essential element} and \emph{something amazing}. The number of such phrases decreased in each
  dictionary-construction iteration, which allows us to assume that this set is relatively small. Nonetheless, we are
  aware that manual construction of this set doesn't take evolution of the language's vocabulary into account.
\end{enumerate}

\section{Future work}\label{s:future}
Plans for future development include dealing with issues detected in above-mentioned experiments. The problem of
detecting implicit references to earlier parts of text is known in natural language processing as coreference
resolution and constitutes an independent field of research as described in 
\cite[p. 614]{d60b7f305e6940309e19a9c288d7284d} or specifically for Polish: \cite{ogro:etal:14}. It is planned to adapt
selected coreference resolution methods to our BigData environment and verify their effectiveness in increasing precision
of our extraction method. 

We plan to achieve better detection of phrase begin/ending points by replacing
construction rules described in section \ref{ss:construction_rules} with Conditional Random Field classifier trained on
sentences scored in our experiment with manually annotated proper phrase boundaries. 
Creating of such golden standard set of sentences with IS-A relations is of course more time consuming than the approach
proposed in this paper.
In case of Hearst patterns it turned out to be a necessity. 
Sentences with Hearst-like enumerations contain more complicated dependency structures which are harder to parse
correctly.

Better catch-all phrases elimination can be done as a post-processing step. Membership in these classes should be
uniformly distributed over instances and subclasses in the taxonomy, so there should be no significant correlation
between membership in these classes and proper classes. Filtering methods based on such correlation will be
investigated.

Taking into account the number of filtered out IS-A relations (starting from support level $2$) it is worthwhile to consider development of other ways of 
assessing their correctness.   
The support level criterion (frequency based) effectively increases quality of extracted information, but at the same time significantly reduces its quantity. 
It would be interesting to choose one of the most popular classification methods (ea. Support Vector Machine or Random
Forest classifier) and check its ability to learn a more sophisticated filtering criterion of incorrect IS-A relations. 
The feature space for this classification problem could be much richer than simple information about occurrence
frequency.
One can use more sophisticated characteristics of IS-A relation like for example: size of class and instance phrase (count in number of words), 
type of sources (nominative, instrumental), popularity of instance and class phrase independently (expressed in number of occurrences among all extracted  
IS-A relations).

It would be also interesting to compare precision of Hearst patterns implemented with pseudo-subclass boosting.

\section{Conclusions}
This paper presents a novel method of IS-A relation extraction from patterns for Polish that is 
different from so popular Hearst patterns and is applicable in inflected languages with free word order. 
Thanks to this method we were able to extract knowledge that may not be expressed in
enumeration constructs defined by Hearst.
Additionally, a method for boosting relation extractions count is introduced. As mentioned at the beginning, thanks
to its simplicity it has potential application in any pattern-based IS-A relation extraction method. 
As experiments showed, the algorithm achieves satisfactory precision 
\footnote{60-80\% precision seems to be achieved by other researchers too, see e.g.  
\cite{Ryu2007AutomaticAO} fig. 4 or \cite{DR04a} table 5.}
(although there is still room for improvement) and is capable of generating high number of taxonomical relations. This
makes it a valuable input source of data for any taxonomy induction task.

It is needless to say that experiments described in this paper do not provide a full statistical overview of millions of
IS-A relations extracted from the corpus of Polish Internet documents. We focus on an assessment of precision of the
proposed IS-A relation extraction method. In-depth statistical analysis of such a dataset is desirable and remains as a
task to be accomplished in the next publication devoted to the research path outlined in the previous section.


\bibliographystyle{IEEEtran}
\bibliography{relation-extraction}

\end{document}